\documentclass{article}

\usepackage{spconf}
\usepackage{amsmath}
\usepackage{cite}
\usepackage[pdftex]{graphicx}
\graphicspath{{./images/}}
\DeclareGraphicsExtensions{.pdf,.png}
\usepackage{amssymb}
\usepackage{mathtools}
\usepackage{xfrac}
\usepackage{nicefrac}
\usepackage{booktabs}
\usepackage{url}
\usepackage{algorithm}
\usepackage{algpseudocode}
\usepackage{framed}
\algrenewcommand{\algorithmiccomment}[1]{\hfill \textit{#1}}
\usepackage{setspace}

\newcommand{\mat}[1]{\mathbf{#1}}
\newcommand{\norm}[1]{{\lVert{#1}\rVert}}

\newcommand{\R}{\mathbb{R}}
\newcommand{\mydef}{\stackrel{\text{\tiny def}}{=}}

\newcommand{\rshift}[2]{{#1}_{\stackrel{\scriptscriptstyle#2\:}{\longrightarrow}}}
\newcommand{\lshift}[2]{{#1}_{\stackrel{\:\scriptscriptstyle#2}{\longleftarrow}}}

\newcommand{\ushift}[2]{{\stackrel{}{\scriptscriptstyle#2}\uparrow}{#1}}
\newcommand{\dshift}[2]{{\stackrel{\scriptscriptstyle#2}{}\downarrow}{#1}}

\newcommand{\T}{\mathsf{T}}
\newcommand{\diag}{\operatorname{diag}}

\title{Exact multiplicative updates for convolutional $\beta$-NMF in 2D}

\name{Pedro J. Villasana T. and Stanislaw Gorlow}
\address{Dolby Sweden\\
	ATG Sound Technology Research\\
	G\"{a}vlegatan 12\,A, 113\,30 Stockholm, Sweden}

\begin{document}
%
\maketitle
\begin{abstract}
In this paper, we extend the $\beta$-CNMF to two dimensions and derive exact multiplicative updates for its factors. The new updates generalize and correct the nonnegative matrix factor deconvolution previously proposed by Schmidt and M{\o}rup. We show by simulation that the updates lead to a monotonically decreasing $\beta$-divergence in terms of the mean and the standard deviation and that the corresponding convergence curves are consistent across the most common values for $\beta$.
\end{abstract}


\section{Introduction}
\label{sec:intro}

Nonnegative matrix factorization (NMF) finds its application in the area of machine learning and in connection with inverse problems. NMF became popular after Lee and Seung derived multiplicative factor updates that made the additive steps in the direction of the negative gradient obsolete \cite{Lee1999}. In \cite{Lee2001}, Lee and Seung give empirical evidence of convergence of the multiplicative updates to a stationary point, using (a) the squared Euclidean distance and (b) the generalized Kullback--Leibler divergence as the contrast function. The factorization's origins can be traced back to \cite{Paatero1994,Paatero1997}.

A convolutional variant of the factorization based on the Kullback--Leibler divergence is introduced in \cite{Smaragdis2004}. There, the idea is to model temporal relations in the neighborhood of a point in the time-frequency plane. The corresponding factor updates are taken from \cite{Lee2001} and lead to a biased factorization. In \cite{Smaragdis2007}, to provide a remedy, multiple coefficient matrices are updated (one for each translation) and the final update is by taking the average over all coefficient matrices. The exact same principles are applied in \cite{Wang2009} to derive a convolutional NMF  based on the squared Euclidean distance. There, the authors combine the updates from \cite{Lee2001} with the averaging from \cite{Smaragdis2007} in an efficient manner. Why these updates are inexact is explained in \cite{Villasana2018_arXiv}. A nonnegative matrix factor deconvolution in 2D based on (a) the squared Euclidean distance and (b) the Kullback--Leibler divergence is found in \cite{Schmidt2006}. It should be pointed out that the update rule for the coefficient matrix is different from those in \cite{Smaragdis2004, Smaragdis2007, Wang2009}. A convolutional NMF has been deployed with arguable success to extract sound objects \cite{Smaragdis2004}, to separate speakers \cite{Smaragdis2007}, to detect onsets \cite{Wang2009}, to automatically transcribe music \cite{Schmidt2006}, and more recently to enhance speech \cite{Sun2015} or to discover recurrent patterns in neural data \cite{Mackevicius2018}.

In this manuscript, we extend our previous work on the $\beta$-CNMF \cite{Villasana2018_arXiv} to two dimensions and derive exact multiplicative updates for its factors. The updates generalize and correct the factor deconvolution proposed in \cite{Schmidt2006}. We further show that the updates lead to a monotonically decreasing $\beta$-divergence \cite{Basu1998} in terms of the mean and the standard deviation and that the corresponding convergence curves are consistent across the most common values for $\beta$. 

\section{Nonnegative matrix factorization}
\label{sec:nmf}

Nonnegative matrix factorization (NMF) is an umbrella term for a low-rank matrix approximation of the form 
\begin{equation}
\mat{V} \simeq \mat{W} \, \mat{H} = \mat{U}
\label{eq:nmf}
\end{equation}
with $\mat{V} \in \R_{\geqslant 0}^{K \times N}$, $\mat{W} \in \R_{\geqslant 0}^{K \times I}$, and $\mat{H} \in \R_{\geqslant 0}^{I \times N}$, where $I$ is the predetermined rank of the factorization. The letters above help distinguish between visible ($v$) and hidden variables ($h$) that are put in relation through weights ($w$). The factorization is usually formulated as a convex minimization problem with an associated cost function $C$ according to 
\begin{equation}
\underset{\mat{W},\,\mat{H}}{\text{minimize}}\ {C{\left(\mat{W}, \mat{H}\right)}} \qquad \text{subject to}\ w_{ki}, h_{in} \geqslant 0 
\end{equation}
with 
\begin{equation}
C{\left(\mat{W}, \mat{H}\right)} \equiv L{\left(\mat{V}, \mat{U}\right)} \text{,}
\label{eq:cost}
\end{equation}
where $L$ is a loss function that assesses the error between $\mat{V}$ and its low-rank approximation $\mat{U}$.

\subsection{$\beta$-divergence}
\label{sec:beta_div}

The loss from \eqref{eq:cost} can be expressed by means of a contrast or distance function between the elements of $\mat{V}$ and $\mat{U}$. So, due to its robustness with respect to outliers for certain values of the input parameter $\beta \in \R$, we resort to the $\beta$-divergence \cite{Basu1998} as a subclass of the Bregman divergence \cite{Bregman1967, Cichocki2010}, which for the points $p$ and $q$ in a closed convex set is given by \cite{Cichocki2010} 
\begin{equation}
d_\beta{\left(p, q\right)} = \begin{dcases}
\frac{p^\beta - q^\beta}{\beta\,{\left(\beta - 1\right)}} - \frac{p - q}{\beta - 1}\,q^{\beta - 1} \text{,} & \beta \not\in {\left\{0, 1\right\}} \text{,} \\
p \, \log\frac{p}{q} - p + q \text{,} & \beta = 1 \text{,} \\
\frac{p}{q} - \log\frac{p}{q} - 1 \text{,}  & \beta = 0 \text{.} 
\end{dcases}
\label{eq:beta_div}
\end{equation}
Accordingly, the $\beta$-divergence between two matrices, $\mat{V}$ and $\mat{U}$, is defined entrywise as 
\begin{subequations}
\begin{equation}
D_\beta{\left(\mat{V} \parallel \mat{U}\right)} \mydef \sum_{k=1}^K\sum_{n=1}^N{d_\beta{\left(v_{kn}, u_{kn}\right)}}
\end{equation}
with
\begin{equation}
u_{kn} = \sum\nolimits_i{w_{ki} \, h_{in}} \text{.}
\label{eq:inner_product}
\end{equation}
\label{eq:entrywise_divergence}%
\end{subequations}
Note that the $\beta$-divergence has a single global minimum for $\sum_n v_{kn} = \sum_{i, n} w_{ki} \, h_{in}$, $\forall k$, even though strict convexity is granted only for $\beta \in {\left[1, 2\right]}$ \cite{Cichocki2010, Fevotte2011}.

\subsection{Multiplicative updates}
\label{sec:mu}

Given that \eqref{eq:beta_div} is continuously differentiable and that the first derivative is monotonically decreasing or increasing if $q < p$ or $q > p$, respectively, we can use gradient descent to find the minimum of \eqref{eq:entrywise_divergence}. Holding $\mat{W}$ or $\mat{H}$ fixed, the iterative update of the variable factor $\mat{X}$ at interation $t$ reads
\begin{equation}
\mat{X}^{t + 1} = \mat{X}^t - \mu \, \nabla C{\left(\mat{X}^t, \cdot^t\right)} \text{,} \quad t \geqslant 0 \text{.}
\label{eq:gradient_descent}
\end{equation}
Splitting the gradient in components with opposite signs,
\begin{equation}
\nabla C{\left(\mat{X}^t, \cdot^t\right)} = \nabla C_+{\left(\mat{X}^t, \cdot^t\right)} - \nabla C_-{\left(\mat{X}^t, \cdot^t\right)} \text{,}
\label{eq:gradient_split}
\end{equation}
and extending the step size $\mu$ to a matrix that changes with $t$,
\begin{equation}
\mu^t \mydef \mat{X}^t \circ {\left[\nabla C_+{\left(\mat{X}^t, \cdot^t\right)}\right]}^{\circ{}-1} \text{,}
\label{eq:mu}
\end{equation}
\eqref{eq:gradient_descent} can be converted to a multiplicative form \cite{Lee1999, Lee2001}:
\begin{equation}
\mat{X}^{t + 1} = \mat{X}^t \circ {\left[\nabla C_+{\left(\mat{X}^t, \cdot^t\right)}\right]}^{\circ{}-1} \circ \nabla C_-{\left(\mat{X}^t, \cdot^t\right)} \text{,}
\label{eq:mur}
\end{equation}
where $\circ$ denotes the Hadamard, i.e.\ entry-wise product, and $\cdot^{\circ{-1}}$ stands for the entry-wise inverse. Multiplicative updates have a faster convergence rate than their additive counterpart.

\subsection{Discrete convolution in 2D}
\label{sec:2d_conv}

As can be seen from \eqref{eq:inner_product}, the weight $w_{ki}$ for the $i$th variable $h_i$ in column $n$ is applied using the scalar product. Should $h_i$ evolve with $n$, we can assume that the current state (or value) of $h_i$ is correlated with its past and future states. We can take this into account by replacing the scalar product in our model by a convolution. Postulating causality and letting the weight $w_{ki}$ have finite support of cardinality $M$, convolution along $n$ writes
\begin{subequations}
\begin{equation}
\sum_{m = 0}^{M - 1}{w_{kim} \, h_{i, n - m}} \mydef {\left(\mat{w}_{ki} \ast \mat{h}_i\right)}_n
\end{equation}
with
\begin{equation}
\mat{w}_{ki} = {\begin{bmatrix} w_{ki, 0} & w_{ki, 1} & \cdots & w_{ki, M - 1}\end{bmatrix}}
\end{equation}
and
\begin{equation}
\mat{h}_i = {\begin{bmatrix} h_{i, n} & h_{i, n - 1} & \cdots & h_{i, n - M + 1} \end{bmatrix}} \text{.}
\end{equation}
\label{eq:conv}%
\end{subequations}
The operation can be converted to a matrix multiplication by lining up the states $\mat{h}_i^\T$ for $n = 0, 1, \dots, N - 1$ in a truncated Toeplitz matrix:
\begin{equation}
\mat{H}_i = {\begin{bmatrix}
h_{i, 0} & h_{i, 1} & \cdots & h_{i, N - 1} \\
0 & h_{i, 0} & \cdots & h_{i, N - 2} \\
\vdots & \vdots & \ddots & \vdots \\
0 & 0 & \cdots & h_{i, N - M}
\end{bmatrix}} \text{.}
\label{eq:toeplitz}
\end{equation}
Using \eqref{eq:conv} and \eqref{eq:toeplitz}, $\mat{V}$ can now be approximated as
\begin{subequations}
\begin{equation}
\mat{U} = \sum_{i = 1}^I{{\left(\mat{W}_i \ast \mat{h}_i\right)}_n} = \sum_{i = 1}^I{\mat{W}_i \, \mat{H}_i}
\end{equation}
with
\begin{equation}
\mat{W}_i = {\begin{bmatrix} \mat{w}_{1, i}^\T & \mat{w}_{2, i}^\T & \cdots & \mat{w}_{K, i}^\T \end{bmatrix}}^\T \text{.}
\end{equation}
\label{eq:conv_model}%
\end{subequations}
In practice, $I$ can be quite large and $M$ is usually small. It is therefore convenient to rewrite \eqref{eq:conv_model} as, see \cite{Smaragdis2004, Smaragdis2007}:
\begin{equation}
\mat{U} = \sum_{m = 0}^{M - 1}{\mat{W}_m \, \rshift{\mat{H}}{m}} \qquad \text{with}\ \mat{W}_m = {\begin{bmatrix} w_{ki\cdot} \end{bmatrix}}_m \text{,}
\label{eq:cnmf}
\end{equation}
where $\rshift{\cdot}{m}$ is a column-wise right-shift operation (similar to a logical shift in programming languages) that shifts all the columns of $\mat{H}$ by $m$ positions to the right, and fills the vacant positions with zeros. The operation is size-preserving. It can be seen that the convolutional NMF (CNMF) has $M$ times as many weights as \eqref{eq:nmf}, whereas the number of hidden variables is equal.

The convolution can be augmented by another dimension \cite{Schmidt2006}, which can be formulated as
\begin{subequations}
\begin{equation}
\sum_{l = 0}^{L - 1}\sum_{m = 0}^{M - 1}{w_{k - l, im} \, h_{li, n - m}} \mydef {\left(\mat{W}_i \ast {\hspace{-.5ex}} \ast \mat{H}_i\right)}_{kn}
\end{equation}
with
\begin{equation}
\mat{W}_{i} = {\begin{bmatrix}
w_{k,i, 0} & \cdots & w_{k, i, M - 1} \\
w_{k - 1, i, 0} & \cdots & w_{k - 1, i, M - 1} \\
\vdots & \ddots & \vdots \\
w_{k - L + 1, i, 0} & \cdots & w_{l - L +1, i, M - 1}
\end{bmatrix}}
\end{equation}
and
\begin{equation}
\mat{H}_i = {\begin{bmatrix}
h_{0, i, n} & \cdots & h_{0, i, n - M + 1} \\
h_{1, i, n} & \cdots & h_{1, i, n - M + 1} \\
\vdots & \ddots & \vdots \\
h_{L - 1, i, n} & \cdots & h_{L - 1, i, n - M + 1}
\end{bmatrix}} \text{.}
\end{equation}
\label{eq:conv_2d}%
\end{subequations}
Using the notation from \eqref{eq:cnmf}, the convolutional data model for \eqref{eq:conv_2d} in two dimensions can be written as
\begin{equation}
\mat{U} = \sum_{l = 0}^{L - 1}\sum_{m = 0}^{M - 1}{\dshift{\mat{W}_m}{l} \, \rshift{\mat{H}_l}{m}} \quad \text{with}\ \mat{H}_l = {\begin{bmatrix} h_{\cdot in} \end{bmatrix}}_l
\label{eq:cnmf_2d}
\end{equation}
and $\mat{W}_m$ as in \eqref{eq:cnmf}. From \eqref{eq:cnmf_2d} one can see that the CNMF in two dimensions has $L$ times as many hidden variables as \eqref{eq:cnmf}. Analogous to the right-shift operator, $\dshift{\cdot}{l}$ is a row-wise down- shift operator.

\subsection{Uniqueness and normalization}
\label{sec:uniqueness}

It is understood that the factorization is not unique. This can be shown easily by the equivalence
\begin{equation}
\mat{U} \equiv \sum_{l = 0}^{L - 1}\sum_{m = 0}^{M - 1}{\dshift{\mat{W}_m}{l} \, \mat{B} \, \mat{B}^{-1} \, \rshift{\mat{H}_l}{m}}
\end{equation}
with $\mat{W}_m \leftarrow \mat{W}_m \, \mat{B}$ and $\mat{H}_l \leftarrow \mat{B}^{-1} \, \mat{H}_l$, for any $\mat{B} \in \R^{I \times I}$ that has an inverse. Nonnegativity still holds for $\mat{W}_m$ and $\mat{H}_l$ if $\mat{B}$ is a nonnegative diagonal matrix. The property is usually used to enforce the same $p$-norm on the matrices $\left\{\mat{W}_i\right\}$:
\begin{equation}
\mat{B} = \diag{\begin{pmatrix} \norm{\mat{W}_1}_p^{-1}, \norm{\mat{W}_2}_p^{-1}, \dots, \norm{\mat{W}_I}_p^{-1} \end{pmatrix}}
\label{eq:normalization}
\end{equation}
with
\begin{equation}
\norm{\mat{W}_i}_p \mydef {\left(\sum_{k=1}^K\sum_{m=1}^M {\begin{bmatrix} w_{\cdot i\cdot} \end{bmatrix}}_{km}^p\right)}^{\nicefrac{1}{p}} \text{.} 
\end{equation} 

\section{$\beta$-CNMF in 2D}
\label{sec:beta_cnmf_2d}

Following up the considerations from Section~\ref{sec:nmf}, we adopt the data model of the CNMF from \eqref{eq:cnmf_2d} and derive multiplicative updates for gradient descent according to \cite{Villasana2018_arXiv} with the entry- wise $\beta$-divergence from \eqref{eq:entrywise_divergence} as the loss function. The result is a $\beta$-CNMF \cite{Villasana2018_arXiv} in two dimensions. A summary follows.

With $u_{kn} = \sum\nolimits_{l, i, m}{w_{k - l, im} \, h_{li, n - m}}$, $p \in {\left\{1, 2, \dots, K\right\}}$, $q \in {\left\{1, 2, \dots, I\right\}}$, and $r \in {\left\{0, 1, \dots, M - 1\right\}}$:
\begin{align}
&\frac{\partial D_\beta{\left(\mat{V} \parallel \mat{U}\right)}}{\partial w_{pqr}} = \sum\nolimits_{k, n}{\frac{\partial d_{\beta}{\left(v_{kn}, u_{kn}\right)}}{\partial u_{kn}}} \cdot \frac{\partial u_{kn}}{\partial w_{pqr}} \nonumber \\
&\quad{} = \sum\nolimits_{k, n}{\left(u_{kn}^{\beta - 1} - v_{kn} \, u_{kn}^{\beta - 2}\right)} \sum\nolimits_{l}{\delta{\left(k - l - p\right)} \, h_{lq, n - r}} \nonumber \\
&\quad{} = \sum\nolimits_{l, n}{{\left(u^{\beta - 1}_{l + p, n} - v_{l + p, n} \, u_{l + p, n}^{\beta - 2}\right)} \, h_{lq, n - r} } \label{eq:derivation} \text{,} 
\end{align}
where $\delta\cdot$ is the Dirac delta function. Choosing $\mu$ in \eqref{eq:gradient_descent} as \eqref{eq:mu} and using \eqref{eq:derivation} in \eqref{eq:mur} leads to the update rule for $\mat{W}_m$:
\begin{align}
\mat{W}_m^{t + 1} &= \mat{W}_m^t \circ {\left[{\sum\nolimits_l{\ushift{\mat{U}^t}{l}^{\circ{\left(\beta - 1\right)}} \, \rshift{\mat{H}_l^t}{m}^{\T}}}\right]}^{\circ{-1}} \nonumber \\ 
&\qquad{} \circ \sum\nolimits_l{{\left[{\ushift{\mat{V}}{l} \circ \ushift{\mat{U}^t}{l}^{\circ{\left(\beta - 2\right)}}}\right]} \, \rshift{\mat{H}_l^t}{m}^{\T}} \text{,} \label{eq:mur_wm}
\end{align}
where $\ushift{\cdot}{l}$ is the up-shift operator. The update rule for $\mat{H}_l$ can be derived in similar fashion \cite{Villasana2018}, resulting in
\begin{align}
\mat{H}_l^{t + 1} &= \mat{H}_l^t \circ {\left[{\sum\nolimits_m{\dshift{\mat{W}_m^t}{l}^{\T} \, \lshift{\mat{U}^t}{m}^{\circ{\left(\beta - 1\right)}}}}\right]}^{\circ{-1}} \nonumber \\ 
&\qquad{} \circ \sum\nolimits_m{\dshift{\mat{W}_m^t}{l}^{\T} \, {\left[{\lshift{\mat{V}}{m} \circ \lshift{\mat{U}^t}{m}^{\circ{\left(\beta - 2\right)}}}\right]}} \text{,} \label{eq:mur_hl}
\end{align}
where $\lshift{\cdot}{m}$ is the left-shift operator, respectively. Algorithm~\ref{alg:betacnmf2} gives a summary of the main processing steps.

\begin{algorithm}
\small
\begin{spacing}{1.1}
\begin{algorithmic}[1]
\Require $v_{kn} \geqslant 0$, $w_{kim}^{t=0} > 0$, $h_{lin}^{t=0} > 0$, $\beta \in {\left[0, 2\right]}$
\Ensure $\mat{U} \simeq \mat{V}$ s.t.\ $u_{kn} \geqslant 0$
\For{$t \gets 1, T$}
	\State $\mat{U} \gets \sum_{l = 0}^{L - 1}\sum_{m = 0}^{M - 1}{\dshift{\mat{W}_m}{l} \, \rshift{\mat{H}_l}{m}}$
	\State $C \gets D_\beta{\left(\mat{V} \parallel \mat{U}\right)}$
	\If{$C < \epsilon$}
		\State \Return
	\EndIf
	\For{$m \gets 0, M - 1$}
		\State $\mat{W}_m \gets \mat{W}_m \circ {\left[{\sum\nolimits_l{\ushift{\mat{U}}{l}^{\circ{\left(\beta - 1\right)}} \, \rshift{\mat{H}_l}{m}^{\T}}}\right]}^{\circ{-1}}$
		\State $\qquad {} \circ \sum\nolimits_l{{\left[{\ushift{\mat{V}}{l} \circ \ushift{\mat{U}}{l}^{\circ{\left(\beta - 2\right)}}}\right]} \, \rshift{\mat{H}_l}{m}^{\T}}$
	\EndFor
	\State $\mat{U} \gets \sum_{l = 0}^{L - 1}\sum_{m = 0}^{M - 1}{\dshift{\mat{W}_m}{l} \, \rshift{\mat{H}_l}{m}}$
	\For{$l \gets 0, L- 1$}
		\State $\mat{H}_l \gets \mat{H}_l \circ {\left[{\sum\nolimits_m{\dshift{\mat{W}_m}{l}^{\T} \, \lshift{\mat{U}}{m}^{\circ{\left(\beta - 1\right)}}}}\right]}^{\circ{-1}}$ 
		\State $\qquad {} \circ \sum\nolimits_m{\dshift{\mat{W}_m}{l}^{\T} \, {\left[{\lshift{\mat{V}}{m} \circ \lshift{\mat{U}}{m}^{\circ{\left(\beta - 2\right)}}}\right]}}$
	\EndFor
\EndFor
\end{algorithmic}
\end{spacing}
\caption{$\beta$-CNMF in 2D}
\label{alg:betacnmf2}
\end{algorithm}

In \cite{Schmidt2006}, multiplicative updates are given for a CNMF in 2D (time and frequency) with the (generalized) Kullback--Leibler divergence and the squared Euclidean distance as the loss or cost function. In the dimension of time, the updates are very much the same as our updates for $\beta = 2$. For $\beta = 1$, there is the minor difference that the $\mat{U}$-matrix in the first line of \eqref{eq:mur_wm} and \eqref{eq:mur_hl} is not shifted, neither up nor to the left. 

\begin{figure*}[!ht]
\centering
\includegraphics[width=\textwidth]{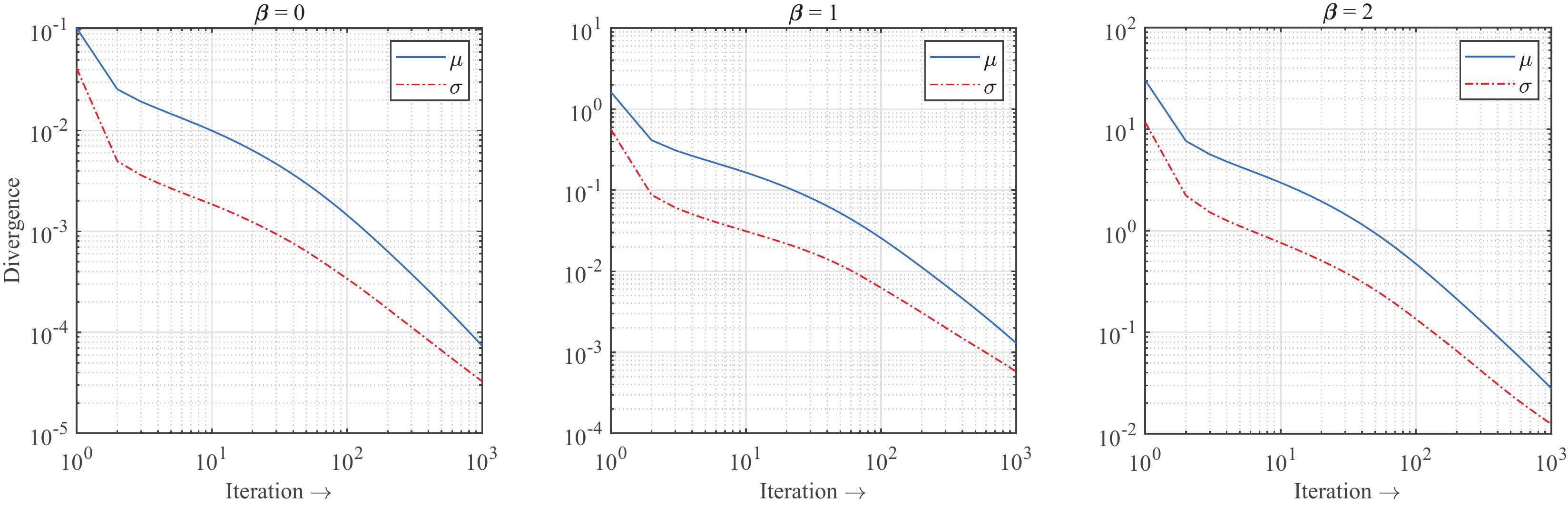}
\caption{Simulation results showing the mean and the standard deviation of the divergence between $\mat{V}$ and $\mat{U}$.}
\label{fig:sim}
\end{figure*}

\section{Simulation}
\label{sec:simulation}

In this section, we simulate and assess the convergence of the newly derived updates for $1 \times 10^3$ iterations. To that end, we generate $1 \times 10^2$ distinct $\mat{V}$-matrices from $M$ $\chi^2$-distributed $\mat{W}_m$-matrices,
\begin{equation}
w_{kim} = \sum_{p = 1}^2{w_{kimp}^2} \sim \chi^2_2 \qquad w_{kimp} \sim \mathcal{N}{\left(0, 1\right)} \text{,}
\end{equation}
and $L$ uniformly distributed $\mat{H}_l$-matrices,
\begin{equation}
h_{lin} \sim \mathcal{U}{\left(0, 1\right)} \text{.}
\end{equation}
We select $M = L = 2$. The factorization is repeated $1 \times 10^1$ times, using random initializations of $\left\{\mat{W}_m^{t = 0}\right\}$ and $\left\{\mat{H}_l^{t = 0}\right\}$ with non-zero entries. So, the curves in Fig.~\ref{fig:sim} were computed over ensembles of $1 \times 10^3$ costs at each iteration (step). The number of visible variables and observations is $K = 1 \times 10^1$ and $N = 2.5 \times 10^1$, while the number of hidden variables $I$ is $5 \times 10^0$. 

As can be seen from Fig.~\ref{fig:sim}, the multiplicative updates are stable (the entry-wise divergence is monotonically decreasing w.r.t.\ both the mean and the standard deviation) and they also are consistent across different values of $\beta$. The difference in scale is because
\begin{equation}
d_\beta{\left(p, q\right)} \equiv p^\beta \, d_\beta{\left(1, \frac{q}{p}\right)} \text{,}
\end{equation}
which evinces that only the Itakura--Saito divergence ($\beta = 0$) is scale invariant. In addition, we measured the run time as a function of the $\beta$-value on an Intel Xeon E5-2637 v3 CPU at 3.5 GHz with 16 GB of RAM. For $\beta = 0$, one iteration takes about 1.41 times longer than for $\beta = 2$, whereas for $\beta = 1$ an iteration takes only a factor of 1.05 longer. The convergence curves have a similar trajectory for different values of $K$, $N$, and $I$.

\section{Conclusion}
\label{sec:conclusion}

In summary, this paper extends our previous work on the $\beta$-CNMF to two dimensions. The $\beta$-CNMF in 2D corrects and generalizes the (2D) nonnegative matrix factor deconvolution by Schmidt and M{\o}rup. It is shown that the new updates are stable and that their convergence behavior is consistent.

\vfill
\pagebreak

\bibliographystyle{IEEEtran}
\bibliography{refs}

\end{document}